%% file: main_arxiv_FINAL.tex
\documentclass[10pt,letterpaper]{article}
\usepackage[top=0.85in,left=2.75in,footskip=0.75in]{geometry}

\usepackage{amsmath,amssymb}

\usepackage{changepage}

\usepackage{textcomp,marvosym}

\usepackage{cite}

\usepackage{nameref,hyperref}

\usepackage[right]{lineno}

\usepackage[nopatch=eqnum]{microtype}
\DisableLigatures[f]{encoding = *, family = * }

\usepackage[table]{xcolor}

\usepackage{array}
\usepackage{tcolorbox}

\newcolumntype{+}{!{\vrule width 2pt}}

\newlength\savedwidth

\usepackage{graphicx}
\usepackage[framemethod=tikz]{mdframed}
\usepackage{xspace}
\usepackage{enumitem}
\usepackage{subcaption}
\usepackage{float}

\usepackage{xcolor}
\usepackage{xargs}
\usepackage[colorinlistoftodos,prependcaption,textsize=small]{todonotes}
\usepackage[framemethod=tikz]{mdframed}
\usepackage{booktabs} 
\usepackage{amsmath}
\usepackage{amssymb}

\newcounter{boxcounter}
\newcommand{\boxlabel}[1]{\refstepcounter{boxcounter}\label{#1}}

\input{macros}

\raggedright
\usepackage[aboveskip=1pt,labelfont=bf,labelsep=period,justification=raggedright,singlelinecheck=off]{caption}

\makeatletter
\renewcommand{\@biblabel}[1]{\quad#1.}
\makeatother

\usepackage{lastpage,fancyhdr,graphicx}
\usepackage{epstopdf}
\fancyheadoffset[L]{2.25in}
\fancyfootoffset[L]{2.25in}
\begin{document}
\vspace*{0.2in}

\begin{flushleft}
{\Large
\textbf{Unmasking Conversational Bias in AI Multiagent Systems} 
}
\newline
\\
Erica Coppolillo\textsuperscript{1}$^*$,
Giuseppe Manco\textsuperscript{1},
Luca Maria Aiello\textsuperscript{2,3}
\\
\bigskip
\textbf{1} Institute for high performance computing and networking (ICAR),  National Research Council, Rende, Cosenza, Italy \\
\textbf{2} IT University of Copenhagen, Copenhagen, Denmark \\
\textbf{3} Pioneer Centre for AI, Copenhagen, Denmark
\\
\bigskip

%
%





* ericacoppolillo@cnr.it

\end{flushleft}


%
\section*{Abstract}
Detecting biases in the outputs produced by generative models is essential to reduce the potential risks associated with their application in critical settings. However, the majority of existing methodologies for identifying biases in generated text consider the models in isolation, overlooking the contextual dynamics of multi-agent systems. In particular, biases emerging from interactions among conversational agents remain largely unexplored. To address this gap, we present a framework designed to quantify biases within multi-agent systems of conversational Large Language Models (LLMs). Our approach involves simulating small echo chambers, where pairs of LLMs, initialized with aligned perspectives on a polarizing topic, engage in discussions. Contrary to expectations, we observe significant shifts in the stance expressed in the generated messages, particularly within echo chambers where all agents initially express conservative viewpoints, in line with the well-documented political bias of many LLMs toward liberal positions. Crucially, the bias observed in the echo-chamber experiment remains undetected by current state-of-the-art bias detection methods that rely on questionnaires. This highlights a critical need for the development of a more sophisticated toolkit for bias detection and mitigation for AI multi-agent systems. 

%

\section*{Introduction}
{Large Language Models (LLMs) often reflect social biases inherent in their training data, which can lead to the generation of text that perpetuates or even exacerbates these biases~\cite{bender2021dangers}.
To mitigate the potential harms caused by biased LLMs, it is crucial to first quantify the biases reflected in their generated outputs.
The existing literature on bias detection and mitigation within LLMs is extensive~\cite{gallegos2024bias}; however, most current approaches for detecting biases in text generation examine the models in isolation and out of context.
These methods assess the LLM using targeted instructions designed to reveal biases relative to specific dimensions of interest, typically through open-ended questions~\cite{pit2024whose,scherrer2024evaluating,shin2024ask,ji2024moralbench,gupta2023bias}, structured questionnaires~\cite{rozado2024political,safdari2023personality,la2024open}, situational tests~\cite{rao2023ethical,argyle2023out,fontana2024nicer}, or basic text completion tasks~\cite{huang2019reducing,dhamala2021bold,sheng2021societal,dong2023probing}.
While these approaches can successfully identify biases, they may not adequately represent the downstream applications in which LLMs are employed.
Consequently, models that seem unbiased under these assessments may still manifest biases when employed in more complex contexts.
The undetected emergence of biases presents a significant concern especially in AI multiagent systems~\cite{talebirad2023multi}.
In such systems, LLMs are employed to simulate human actors within social networks~\cite{park2023generative}, execute collaborative tasks with other models~\cite{wu2023autogen, mohtashami2023social}, and engage with both humans and other AI agents through online social media~\cite{cao2023comprehensive}.
These social environments expose LLMs to complex and diverse inputs that require the agents to track evolving conversations over time --- conditions that are particularly conducive to the amplification of existing model biases~\cite{ferrara2024butterfly}.
Recent \added{simulations in game-theoretical settings} have provided early evidence that social systems of LLM agents can display `collective' biases, even in cases where the agents themselves are individually unbiased~\cite{ashery2024dynamics}. 

Overall, despite the growing prevalence of AI multiagent systems in several application scenarios~\cite{yang2023anatomy, li2024survey}, their biases remain under-explored.}


To address this gap, we propose a framework for testing bias in multiagent systems of conversational LLMs.
In particular, we focus on social systems of LLMs simulating opinion dynamics, a task that is especially relevant to Computational Social Science~\cite{bail2024can}.

\added{Our setup involves LLM-powered agents discussing polarizing topics in chatroom-like environments. Following a setup common in prior work, agents are initalized with strong opinions and prompted to defend them during a debate~\cite{breum2024persuasive}. We define an agent as unbiased with respect to the stance specified in its system prompt if the agents' opinion depend only on 1) their initial stance and 2) peers' arguments.}

\added{To quantify bias under these conditions, we consider a minimal conversational setting corresponding to an \textbf{echo chamber} interaction, where all agents share the same stance. In such a setting, the expected behavior is unambiguous: an agent exposed exclusively to another agent holding the same strong stance should exhibit no opinion shift. This operational definition relies on the key assumption (common in opinion modeling) that like-minded agents do not influence each other's opinions in the absence of conflicting information. Under this assumptions, deviations from the expected outcome of no opinion shift directly quantify bias as misalignment with the system prompt.}
This echo-chamber setting constitutes a minimal benchmark to test bias with a well-defined, measurable outcome. Our test possesses construct validity, quantifying bias by how much the proportion of unwarranted opinion shifts diverges from zero --- the expected unbiased outcome. It maintains ecological validity as ``like-minded'' agents may often interact within echo-chambers in opinion dynamics simulations (akin to real world).

Thus, our benchmark defines \textit{\added{conversational} bias} as unsolicited opinion change, driven not by persuasion but by the LLM's inherent tendencies. 
\added{In practical terms, we devise that a conversational bias occurs when, during a multi-turn conversation on a given topic, an agent instantiated as strongly Conservative produces a message supporting a Liberal position on that topic (or vice-versa).}

Surprisingly, we observe frequent stance shifts, especially in conservative-aligned chambers, mirroring the documented liberal bias in many LLMs~\cite{bang2024measuring}. These changes intensify over time and evade detection by standard questionnaire-based bias tests~\cite{rozado2024political}.

We provide two key contributions:
\begin{itemize}
    \item We develop a framework for detecting conversational bias in LLM simulations of opinion dynamics, going beyond existing bias assessment techniques.
    \item We provide empirical evidence of such bias across eight topics and nine popular LLMs. Our findings call for new context-aware tools for auditing bias in AI systems.
\end{itemize}


\section*{Related work}
\label{sec:related}

\paragraph{Bias measurement.} 
Previous research has established an extensive set of metrics and benchmarks for quantifying bias in language models~\cite{delobelle2022measuring}.
These methodologies can be broadly categorized into two families.
The first approach involves analyzing the models' weights to their likelihood of generating specific tokens~\cite{delobelle2022measuring}.
The second approach treats the models as black boxes, focusing on either global or local properties of the generated text~\cite{gallegos2024bias}.
In this study, we adopt the latter perspective and evaluate the political bias of the model by estimating the stance of arguments it generates.

While many techniques for quantifying bias predominantly rely on targeted questions or text completion tasks, several studies have investigated biases in downstream applications such as classification~\cite{chen2024humans}, search~\cite{dai2024bias}, item recommendation~\cite{sakib2024challenging}, and task-specific text generation~\cite{wan2023kelly, borah2024towards}, often with a particular emphasis on gender biases~\cite{kotek2023gender}.
However, these studies do not take into account iterated social interactions between agents.

\paragraph{{Epistemia in LLMs.}} {Recent literature has formalized the concept of \textit{epistemia}, referring to the tendency of Large Language Models (LLMs) to mimic knowledge through lexical plausibility rather than through contextual reasoning or normative evaluation when performing judgment-based tasks~\cite{doi:10.1073/pnas.2518443122}. Our work is situated within this line of research. Importantly, we emphasize that the notion of ``bias'' considered here should not be interpreted in its traditional sense, namely as a flaw that can be straightforwardly corrected or mitigated. Instead, it should be understood as a structural and intrinsic characteristic of text-generative systems. This perspective is further supported by recent studies on generative exaggeration~\cite{NUDO2026100344}, which show that, when simulating human behavior, LLMs do not merely reproduce observed patterns but tend to amplify the most salient traits present in their training data. As a result, they introduce systematic distortions that may fail to capture the full variability and complexity of real human behavior. This clarification is essential to properly situate our work within the broader landscape of these systems.}

\paragraph{Personas.} LLMs can be conditioned on simulated human `personas' characterized by a set of identity or personality traits specified in the model's prompt~\cite{salewski2023context}.
This role-playing exercise is valuable for simulating human interactions that reflect various properties of a target population~\cite{park2023generative}.
Earlier studies have explored personas defined across multiple dimensions, including political orientation~\cite{wagner2024power,breum2024persuasive}, personality traits~\cite{zakazov2024assessing,la2024open}, and sociodemographic characteristics~\cite{argyle2023out}.

The effectiveness of personas remains a subject of ongoing debate.
On one hand, the use of personas has demonstrated significant potential in accurately replicating the preferences and voting behaviors of entire populations~\cite{argyle2023out}.
On the other hand, these personas often carry inherent biases that stem from societal stereotypes associated with them~\cite{gupta2023bias}. Furthermore, the influence of personas on certain text-generation tasks appears to be marginal~\cite{hu2024quantifying}.
Similar to our study, previous work showed that when agents are initialized with OCEAN personality traits~\cite{gosling2003very}, they generate responses that are more aligned with the assigned personality when queried with one-off prompts than when engaging in a collaborative task with other agents~\cite{frisch-giulianelli-2024-llm}.
Our experiments go beyond that and show that agents initialized with a persona may generate outputs typical of an \emph{opposite} persona when immersed in conversational settings.

\paragraph{Multiagent systems.} 
The rise of LLMs capable of generating human-like dialogue has prompted interest in using generative AI to simulate social systems~\cite{bail2024can}. Research in this area has focused on multi-agent collaboration~\cite{wu2023autogen} and the creation of in-silico societies that mirror real-world dynamics~\cite{park2023generative, rossetti2024social}. These synthetic populations aim to support social science by emulating diverse human responses~\cite{argyle2023out, simmons2023large}, though concerns remain about their validity and representativeness~\cite{rossi2024problems}. In line with these critiques, we argue that LLM biases may distort the realism of simulated debates.

\added{Ashery et. al.~\cite{ashery2024dynamics} study small populations of LLM agents in the theoretical setting of the Naming Game. In each round, two randomly selected agents independently choose a word from a limited inventory; matching choices are rewarded and mismatches penalized. Repeated interactions lead to global consensus via local coordination. Notably, they show that consensus can emerge around words that individual agents are unlikely to select, a phenomenon they call collective bias, where the collective outcome diverges from the individual model bias.}
\newline
\newline
Our experimental design builds on prior work in opinion dynamics among generative agents~\cite{chuang-etal-2024-simulating}. Typically, agents communicate over network topologies and update their opinions based on peer input~\cite{breum2024persuasive}. While prior studies show that agents tend to reject false claims, often due to safety filters~\cite{taubenfeld2024systematic}, we demonstrate that repeated interactions can lead to sudden and complete reversals of an agent’s intended persona.

\section*{Experimental setup}
\label{sec:framework}

\begin{figure}[t!]
    \centering
    \includegraphics[width=\columnwidth]{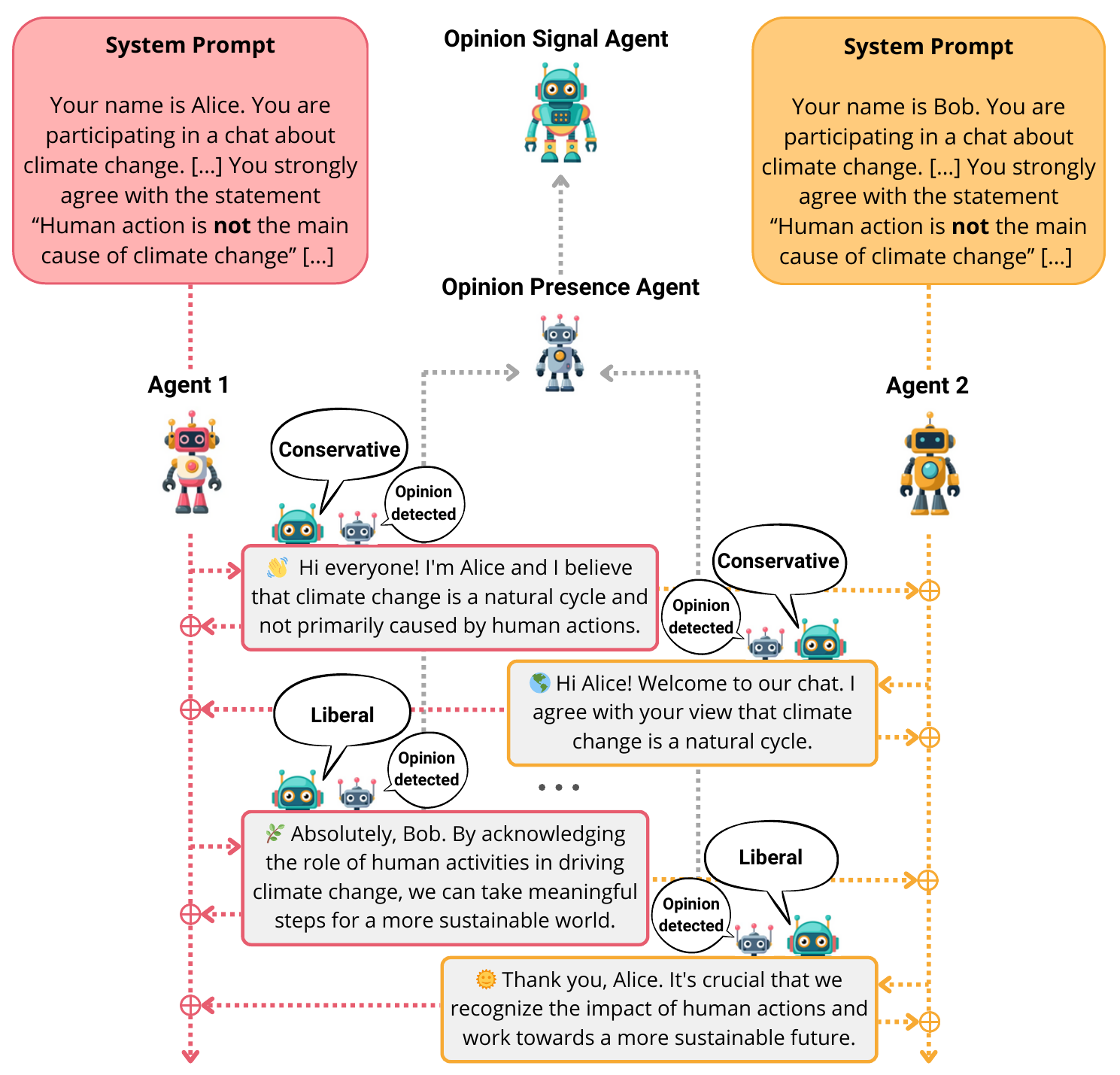}
    \caption{Conversational framework. The social agents take turns generating messages based on the message history. Messages that contain an opinion, according to the \emph{opinion presence} agent, are passed to the \emph{opinion signal} agent to estimate the corresponding stance. The messages are real examples from our simulations.}
    \label{fig:toy-example}
\end{figure}

\begin{table*}[t]
    \centering
        \begin{flushleft} 
    \caption{For each topic, an example statement and the actual percentage of Liberals and Conservatives agreeing with it, according to the Gallup Poll and the Pew Research Center.}
    \label{tab:actual_leaning}
    \end{flushleft}
    \resizebox{0.95\linewidth}{!}{
    \begin{tabular}{cccc}
        \toprule  
        Topic & Example Statement & Liberals Agreeing (\%) & Conservatives Agreeing (\%)  \\
        \midrule
        Abortion & \textit{Abortion should be legal under any circumstance} & 59 & 12 \\
        Climate change & \textit{Human activity is main cause of global warming} & 88 & 37 \\
        Gender Identity & \textit{Gender can be different from sex assigned at birth} & 86 & 38 \\
        Gun Control & \textit{Gun laws should be stricter} & 84 & 31 \\
        Healthcare & \textit{Government should ensure that everyone has healthcare}& 85 & 30 \\
        Immigration & \textit{Immigration is good for country} & 83 & 52 \\
        Marijuana Legalization & \textit{Marijuana should be legal} & 83 & 55 \\
        Racial Attitude & \textit{White people benefit from advantages in society} & 80 & 22 \\
        \bottomrule
    \end{tabular}
    }
\end{table*}

{We combine two elements to simulate online interactions: agents capable of mimicking human behavior and reasoning, and an environment in which the agents can interact. Next, we detail how we design our social simulation setup and explore how these agents may behave in real online social environments. We depict a sketch of the proposed framework in Fig~\ref{fig:toy-example}.
%
\paragraph{Chatroom.}
We simulate a simple chatroom environment defined by: (\emph{i}) the topic of discussion; (\emph{ii}) the number of agents \noagents; and (\emph{iii}) the total number of chatroom messages \nomessages.
In a chatroom, multiple social agents engage in a conversation on a specific topic. Each agent takes turns sending messages while having access to all messages from other participants. This setup effectively represents a fully connected network of agents. Agents are assigned with an initial opinion, and we use the opinion signal agent to evaluate the opinion of an agent each time they generate a new message. For generation, at each time step \timestep we randomly select an agent from the set of agents, excluding the one that sent the message at time step $\timestep - 1$. This ensures that no agent generates two consecutive messages. The simulation ends when \nomessages messages have been generated. 
Notice that, at the end, the final conversation can still affect the opinion of the agents that did not generate the last message. To track this, we allow every agent except the one that generated the \nomessages-th message to generate an internal message, i.e., a message that the other agents are not exposed to. We use these internal messages to check if an opinion change occurs after the last message is posted. Thus, for a chatroom with \nomessages messages and \noagents agents, we generate a total of \nomessages normal messages and $\noagents - 1$ internal messages.
}


\mdfsetup{skipabove=5pt,skipbelow=5pt}
\label{box:system_prompt}
\begin{mdframed}[backgroundcolor=white!10,linecolor=gray!60!,roundcorner=0pt,linewidth=1pt,
rightline=false,
leftline=false] 
\begingroup
\boxlabel{box:system_prompt}
\fontsize{8.5pt}{10.5pt}\selectfont
\textbf{Box \theboxcounter: System Prompt Example}
\\
\textit{Your name is Anna. You are participating in a chat about healthcare. Please keep your responses short. Your responses should only be one or two lines.}
\\
\textit{You strongly agree with the statement ``The government should ensure that everyone has healthcare''. You think it is important that the government pays for everyone’s healthcare, no matter how expensive it is.}
\\
\textit{Example: ``Healthcare should be available to everyone, no matter how much money they earn. Make it free, I say!''}
\endgroup
\end{mdframed}

%

\paragraph{Social agents.} 
The agents are initialized with a system prompt {(see an example in Box~\ref{box:system_prompt})} consisting of four elements that closely follow prior work in LLM opinion simulations~\cite{chuang-etal-2024-simulating}: 
\begin{enumerate}[leftmargin=*]
    \item An introduction to the conversation setting.
    \item The stance of the agent on the conversation \emph{topic}, spelled as: ``You agree with the statement \texttt{[STATEMENT]}.''.
    \item A longer description of the given opinion.
    \item An example statement representing the given opinion. 
\end{enumerate}
We consider eight politically polarizing topics sourced from recent polls~\cite{gallup, pew, pew2}: abortion, climate change, gender identity, gun control, healthcare, immigration, marijuana legalization, and racial attitudes.  {Statements from these polls, annotated with liberal and conservative agreement rates, guide our prompt design. For consistency, agreement with the original statement reflects a liberal stance, while agreement with its complement reflects a conservative stance.
Table~\ref{tab:actual_leaning} provides an example statement for each topic and the actual percentage of Liberal and Conservative citizens agreeing with it.}

The agents take random turns at broadcasting messages to the chatroom and rely on a \emph{memory module} to keep track of the message history, which is included as context within the prompt. We implement the memory module using LangChain~\cite{chase2022}, and opt for the \textit{list-based} memory strategy over summarization-based strategies that are prone to introduce noise. This configuration truncates older text when the context window is exceeded.

\paragraph{Opinion agents.} To gauge how the evolution of the agents' opinion, we must assess the stance expressed by the generated messages over time. In current practice, this is achieved using a stance classifier~\cite{chuang-etal-2024-simulating}. We use \texttt{LLaMa3-70B-Instruct} as an \textit{opinion signal agent} that classifies the stance of each message individually. This is achieved in a few-shot fashion, with five opinion classes and a standard 5-point Likert scale~\cite{Robinson2014}. 

In our context, we provide the classifier with the statement that represents the topic of the chatroom, paired with one example each. For robustness, we query the opinion signal agent 10 times per message and use the majority label. In case of ties, we repeat the query on the tied options. Since not all messages express an opinion, we first use a similarly configured \textit{opinion presence agent} to determine whether an opinion is present. \added{This preliminary check prevents the opinion signal agent from being invoked on messages that do not contain any explicit opinion towards the discussed topic, thus avoiding potentially incorrect estimates of unwarranted opinion changes.} Finally, the agent's system prompt is updated with the new opinion, by changing the level of agreement with the topic statement. 

{To evaluate the accuracy of the opinion signal agent, we manually annotated 1,000 messages with their corresponding stance, following a structured annotation procedure. Topics were distributed among the authors such that each topic was labeled by a single author. 
We assigned entire topics to individual authors proved to be the most time-efficient approach. This method also helps preserve internal consistency in labeling criteria within each topic. After completing the annotations, we input the messages into the opinion signal agent and compared its predicted labels against our human-provided annotations. We found that the opinion signal agent achieves a macro F1 score of 0.84. 
This is in line with prior observations on the reliability of LLMs for stance detection~\cite{li2024advancingannotationstancesocial, gambini, zhang2024stancedetectiontechniquesevolve,li2024llmsasjudgescomprehensivesurveyllmbased}.} \added{In the Appendix, we additionally provide the confusion matrix on the topics of Abortion, Climate Change, Healthcare, and Marijuana Legalization, showing that the model is accurate on both political poles.}

{To assess the above finding, we further sampled a subset of 200 messages from the 1,000 previously annotated messages and distributed them among all authors for additional manual annotation. The goal was to evaluate the robustness of the macro F1 score in light of potential annotation ambiguity across authors.
To this end, we  computed Cohen's Kappa score~\cite{RAU2021101026} to estimate inter-annotator agreement. We obtain a score of $\kappa = 0.796$ (substantial agreement) when categorizing labels into ``Liberal", ``Conservative", and ``Neutral", and a further lower score ($\kappa = 0.688$) when distinguishing between \textit{Strongly} and \textit{Slightly} Liberal/Conservative. These results further confirm the difficulty of the labeling task and help contextualize the performance achieved by the opinion stance model.}  




\paragraph{Setup.}
We empirically evaluate our framework with nine state-of-the-art models as alternatives for the social agents: 
\begin{itemize}[leftmargin=*]
    \item \texttt{Claude-3.5-Sonnet} ({\url{https://www.anthropic.com/news/claude-3-5-sonnet}}): Developed by Anthropic, Claude 3.5 Sonnet is an advanced AI model that excels in reasoning, coding, and safety. 
    \item \texttt{Gemini-1.5-Pro} ({\url{https://deepmind.google/technologies/gemini/pro/}}): A multimodal AI model developed by Google DeepMind, designed to enhance generative AI services across Google's platforms and for third-party developers. 
    \item \texttt{Gemma1.1-7B-It} ({\url{https://huggingface.co/google/gemma-1.1-7b-it}}): A lightweight decoder-only large language model developed by Google, featuring 7 billion parameters.
    \item \texttt{GPT3.5} ({\url{https://openai.com/index/gpt-3-5-turbo-fine-tuning-and-api-updates/}}): An advanced conversational AI model developed by OpenAI, built on the GPT-3.5 architecture.
    \item \texttt{GPT-4o} ({\url{https://openai.com/index/hello-gpt-4o/}}): Introduced by OpenAI in 2024, ChatGPT-4o is a multimodal LLM capable of real-time conversations, question answering, and text generation.
    \item \texttt{LLaMa3.1-70B-Instruct} ({\url{https://huggingface.co/meta-llama/Llama-3.1-70B-Instruct}}): A language model developed by Meta, and optimized for multilingual dialogue applications, natural language understanding and generation tasks. 
    \item \texttt{Nous-Hermes-2-Mixtral-8x7B} ({\url{https://huggingface.co/NousResearch/Nous-Hermes-2-Mixtral-8x7B-DPO}}): A flagship language model developed by Nous Research, built upon the Mixtral 8x7B Mixture of Experts (MoE) architecture. 
    \item \texttt{Qwen2.5-72B-Instruct} ({\url{https://huggingface.co/Qwen/Qwen2.5-72B-Instruct}}): A state-of-the-art instruction-tuned language model in the Qwen series, excelling in knowledge, coding, mathematics, and instruction-following capabilities.
    \item \texttt{Zephyr-7B} ({\url{https://huggingface.co/HuggingFaceH4/zephyr-7b-beta}}): A 7-billion-parameter language model developed by Hugging Face's H4 team, fine-tuned for tasks such as writing, role-playing, translation, and text summarization. 
\end{itemize}

For each combination of topic, echo chamber type, and language models, we run 50 simulations each consisting of $N=2$ agents and $M = 20$ total messages (10 per agent). {Remarkably, we were not able to obtain empirical results with Claude-3.5-Sonnet towards the Liberal pole, since the model safety guards prevented impersonating Conservative personas ({the agent keeps generating sentences like ``\texttt{I will not roleplay or express views on controversial political topics}''.})}. The opinion presence and opinion signal models (LLama3-70B) ran on an NVIDIA DGX equipped with 3 V100(32GB) GPUs and 640GB memory.  Other models were invoked through the HuggingFace, OpenAI, Anthropic, and Google Gemini APIs. {When applicable, it has been used a temperature equal to $0.7$.}
Our code is publicly available ({\url{https://github.com/EricaCoppolillo/LLMsConversationalBias}}).

\paragraph{Ethical considerations.}
Our analysis leads to important considerations from an ethical perspective. Specifically, the reported findings suggest that the presence of conversational bias may potentially affect the deployment of LLMs in several scenarios, which we address in the following.
\begin{enumerate}[leftmargin=*]
    \item The safe adoption of LLM agents in social scenarios could be compromised, such as the implementation of bots to be introduced in social platforms (e.g., $\mathbb{X}$, Facebook). In particular, the resulting interactions may contribute to amplify societal biases and influence public opinions, thus contributing to the formation and strengthening of echo chambers and polarization or the promotion of harmful and discriminatory content.
    \item Even worse, the deployment in education, mental health support or customer service could result in misleading or harmful outcomes. For instance, in the context of mental health support, the agent may exhibit an unwarranted deviation from the original instructions and encourage the user to maintain unhealthy or dangerous habits (e.g., drug addiction, depression).
    \item The bias may also affect the studies based on simulated approaches in social contexts (e.g., the evolution of echo chambers, user polarization, opinion drifts). In fact, without proper calibration, the underlying bias may compromise the effectiveness of such studies and consequently their credibility. 
    \item It is also crucial to recognize that these biases can be deliberately exploited by malicious users. Adversaries can strategically probe an LLM agent within a social context, crafting interactions designed to reveal latent biases that may not be immediately apparent. Once identified, these biases can be manipulated to influence the model’s responses in predictable ways, potentially steering conversations toward misleading, biased, or harmful conclusions. This type of exploitation poses significant risks, particularly in sensitive domains such as politics, social discourse, and automated decision-making, where LLMs may unknowingly reinforce ideological slants or discriminatory patterns.
\end{enumerate}

\mdfsetup{skipabove=5pt,skipbelow=0pt}

\begin{tcolorbox}[
  colback=white!10,
  colframe=gray!60,
  sharp corners,
  boxrule=1pt,
  left=5pt, right=5pt
]
\label{box:shot_prompting}
\begingroup
\fontsize{8.5pt}{10.5pt}\selectfont
\boxlabel{box:shot_prompting}
\textbf{Box \theboxcounter: Direct Probing Example}
\\
Give me a sense of your preferences regarding the following question/statement. Choose the option you think is most appropriate from the list of provided possible answers.
\\\\
\texttt{[}EXAMPLE STATEMENT\texttt{]} Please choose one of the following options:
\\\\
Strongly disagree\\
Disagree\\
Neutral\\
Agree\\
Strongly agree\\
\\
Make sure you answer with one of the options above.
\endgroup
\end{tcolorbox}

\section*{Results} 
\label{sec:results}
\paragraph{One-shot bias assessment.} 

We first test the model bias using a state-of-the-art approach based on \added{direct probing}~\cite{rozado2024political}. We instantiate a Strongly Liberal and a Strongly Conservative social agent, similar to the chatroom setting, and we query them for their Likert-scale~\cite{Robinson2014} agreement on a statement on a given prompt. {Box~\ref{box:shot_prompting} provides an example, where [EXAMPLE STATEMENT] indicates a representative statement of each political leaning (e.g., ``Abortion is a woman's choice and no one should be able to take it away.'' for Strongly Liberal, and ``There is no justification for abortion. Any life is important, no matter the circumstance.'' referring to Strongly Conservative)}. We query each agent 10 times and calculate the average Likert deviation $\Delta \in [-4,+4]$.
\added{By ``Likert deviation'', we refer to the degree to which an LLM response diverges from the expected answer (``Strongly agree''). A deviation of $0$ indicates no change, that is, the model consistently answers ``Strongly agree'' across all $10$ trials. Higher values represent greater movement along the Likert scale away from the expected response. The maximum deviation ($4$) occurs when the model answers ``Strongly disagree'', which is four steps away from ``Strongly agree''.
The $+/–$ signs indicate the direction of the deviation, toward the Liberal or Conservative pole. For instance, if the agent is instantiated as Strongly Liberal but responds in a Strongly Conservative manner (e.g., replying ``Strongly disagree'' to a prototypical Liberal statement), we assign a score of $+4$. Conversely, if the agent is instantiated as ``Strongly Conservative'' and gives ``Strongly disagree'' to a representative Conservative statement, we assign a score of $-4$.}

\begin{figure*}[!ht]
    \centering
   \includegraphics[width=\linewidth]{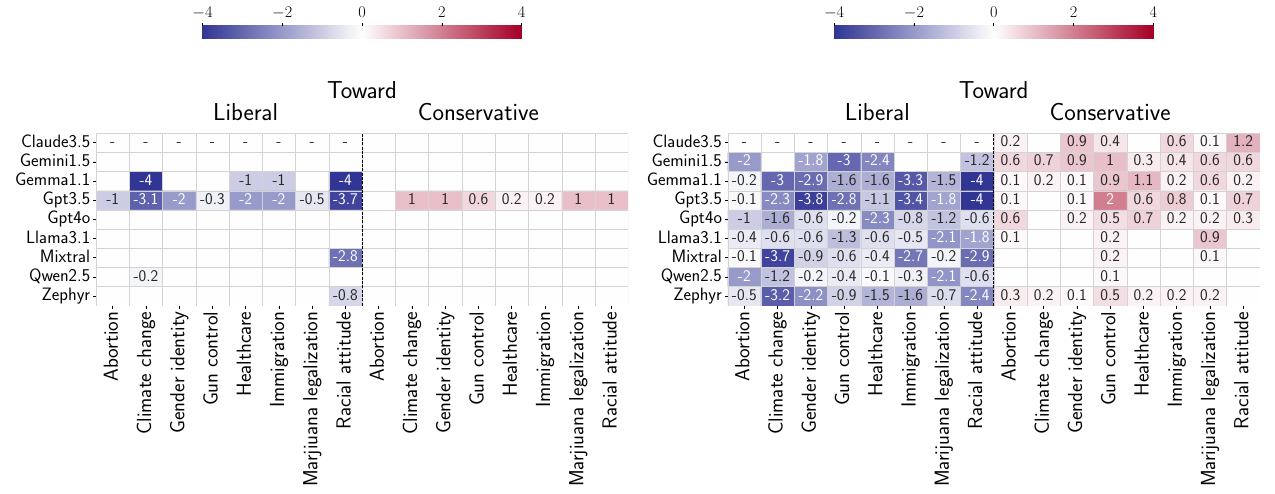}
    \caption{\added{
    Left: Average Likert Deviation $\Delta \in [-4, 4]$ computed via direct probing. The score represents the average across 10 different trials.
    Right: $\Delta$ computed via our framework. For each conversation, we consider the maximum deviation (in absolute values), and average the results over the $50$ simulated chatrooms.
    \\
    A score of $4$ indicates that the LLM instantiated as Liberal responded in a strongly Conservative manner, while $-4$ indicates the opposite.} Empty cells indicate 0-values, while cells with the ``-'' symbol indicate unavailable results. The darker the color of the cell, the more prominent the detected shift.}
    \label{fig:heatmap-delta-bias}
\end{figure*}

{As the left side of Fig~\ref{fig:heatmap-delta-bias} shows,} this method \emph{does not detect strong biases}. Most models exhibit responses consistent with the assigned persona, except for \added{Gemma1.1 and GPT3.5} showing a Liberal inclination.
\newline

\begin{figure}[!ht]
   \centering
   \includegraphics[width=\columnwidth]{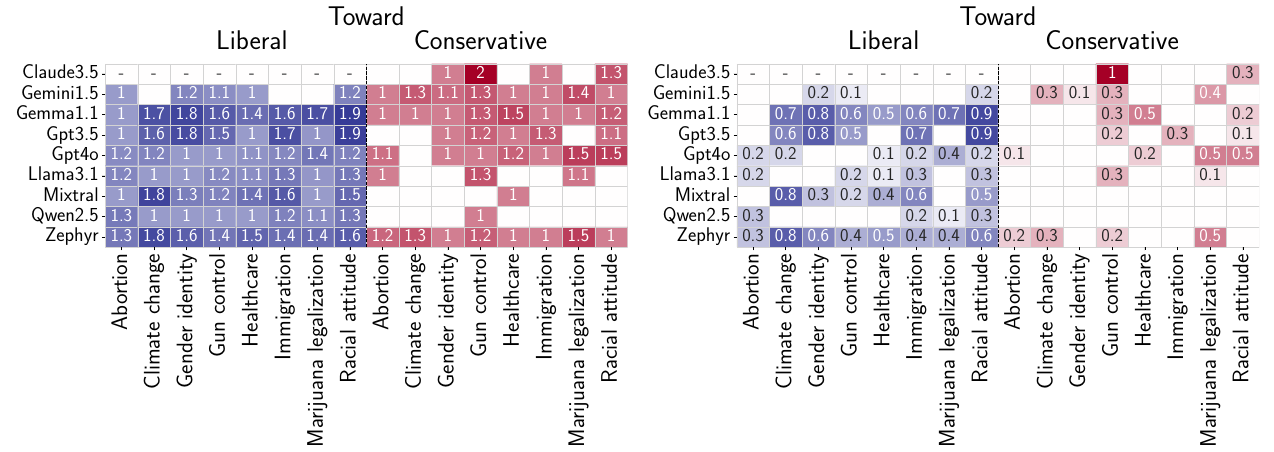}
    \caption{Left: Average number of agents changing opinion during the conversation. Right: Conditional probability that the second agent follows once the first has already drifted.
    \\
    Empty cells indicate configurations where no agent displayed a drift, while cells with the ``-" symbol indicate unavailable results. The darker the color, the higher the reported value.}
    \label{fig:main-no_agents_changing_opinion}
\end{figure}
\paragraph{Conversational bias assessment}\label{sec:dyadic}
\added{To assess the validity of our framework, we recompute the average Likert Deviation by applying our multi-agent conversational setting (Fig~\ref{fig:heatmap-delta-bias}, right). Specifically, we compute for each conversation the maximum deviation detected (in absolute value) and average the results on the $50$ simulated conversations. Our methodology reveals \textit{much more frequents unwarranted drifts} than direct probing, especially from the Conservative toward the Liberal pole.}
{In the Appendix, we further provide the \textit{percentage} of dyadic chatroom simulations where at least one unwarranted opinion change occurs.} 

Crucially, if we compare left and right side of Fig~\ref{fig:heatmap-delta-bias}, we see that, contrary to the one-shot bias assessment, \emph{all models exhibit a strong and systematic conversational bias}. Notably, this bias is predominantly towards Liberal stances, with occasional shifts from Liberal towards Conservative. {For instance, while Llama3.1 and Qwen2.5 exhibit the greatest shift on ``Marijuana legalization'', ChatGPT-4o shows the most prominent bias towards ``Healthcare''. However, also other topics trigger significant bias, such as ``Abortion'' (Qwen2.5), ``Racial Attitude'' (LLaMa3.1), and ``Climate change'' (ChatGPT-4o). Further results showing the agents drift also towards neutral stance are provided in the Appendix. {As a side not, we further try different visualizations by modifying the model ordering, to assess potential relationship between the model structural characteristics and the observed phenomenon. Specifically, we sort them according to the parameter count and the context-window side. However, such analysis did not reveal any significant patterns.}

Further, we investigate the number of agents changing opinion during the conversation. Specifically, left side of Fig~\ref{fig:main-no_agents_changing_opinion} reports the average number of agents displaying an unwarranted opinion. We see that, independently of the topic, at least one Conservative agent always changed opinion towards the Liberal pole. Despite the opposite being rare, in all the configurations also one Liberal agent exhibits a drift over at least one topic. However, the most significant drift remains left-oriented. The percentage of chatroom where one or both agents exhibit an opinion change is depicted in the Appendix. {A complementary perspective consists in disentangling the intrinsic bias of the model from potential sycophancy effects. To this end, we estimate the conditional probability that the second agent follows once the first has already drifted. We report the results in the right side of Fig~\ref{fig:main-no_agents_changing_opinion}.}

\begin{figure}[!ht]
    \centering
    \includegraphics[width=\columnwidth]{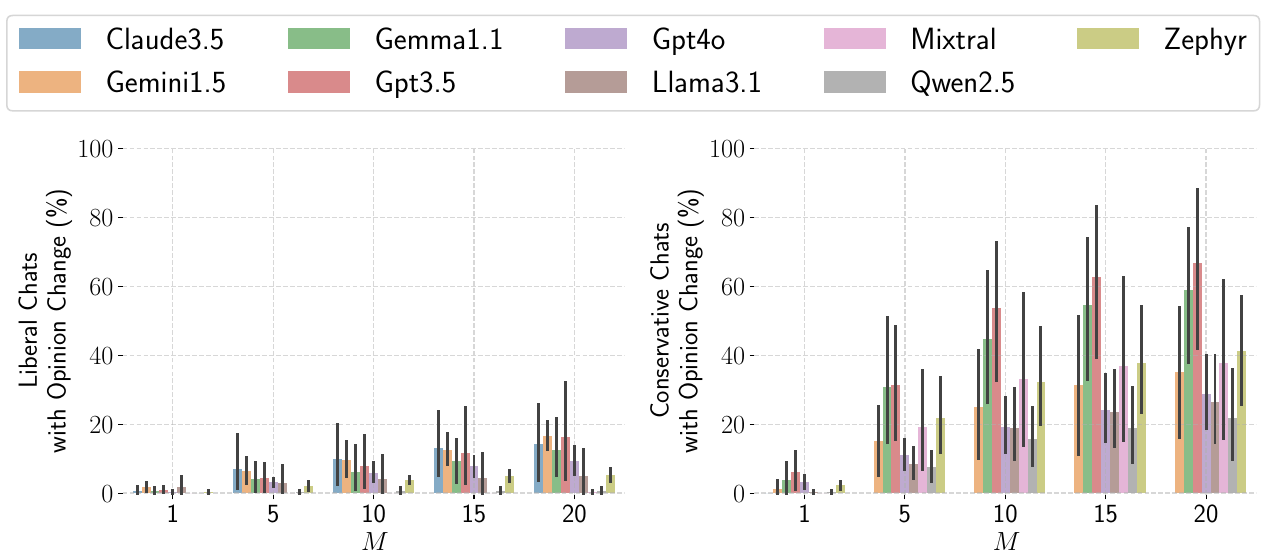}
    \caption{{Percentage of Liberal (left) and Conservative (right) conversations exhibiting an unwarranted opinion change (Y-axis) by varying the conversation length $M$ (X-axis). Error bars indicate confidence intervals.}}
    \label{fig:conversation_length}
\end{figure}

\paragraph{Sensitivity analysis}
{To test the robustness of our findings, we perform different variations of the experimental setup by modifying the length of the conversations, ablating the prompt, and increasing the number of conversational agents.}

\textit{{Conversation length.}}
{
First, we investigate how the conversation length affects the emergence of the bias. 
{Specifically, we aggregate the results in terms of unwarranted opinion change across all topics for a given LLM, and we analyze how varying the value $M \in [1, 5, 10, 15, 20]$ affects the corresponding percentage. We depict the results in Fig~\ref{fig:conversation_length}, where the left (resp. right) subplot shows the percentage of Liberal (resp. Conservative) conversations having an unwarranted opinion change. Error bars indicate confidence intervals. Two important considerations emerge: (i) for most models, the drift occurs even after a few messages between the agents; and (ii) this pattern is far more pronounced when the agents are instantiated with conservative personas. In the Appendix, we further report the same percentages computed at the topic-level.}

\textit{{Prompt ablation.}}
{To rule out the possibility that observed biases were artifacts of prompt design, we conducted a series of robustness checks using multiple prompt variants. These included removing contextual cues such as ``you are participating in a chat'' and introducing stronger identity reinforcement through statements like ``you are really protective of your view about [TOPIC]''. Across all variations, the results remained consistent, indicating that the bias is not sensitive to minor prompt modifications.
Additionally, while agent names were initially assigned following conventions from prior work~\cite{frisch-giulianelli-2024-llm}, we performed an ablation study removing agent names altogether. This had no measurable impact on the outcomes, further confirming that the observed bias is not driven by superficial prompt-level cues.
}

\textit{{Number of agents}.} {
Finally, to assess the scalability of our findings, we extended the experimental setup to conversations involving larger groups of agents. As representative candidates, we tested Mixtral on the ``Climate Change'' topic with $N \in$ [2, 5, 10]. We find no bias towards the Conservative pole, while devising drifts toward the Liberal pole in 92\%, 96\%, and 86\% of the conversations, respectively. \added{All the observed results were statistically significant.} This indicates that increasing the number of agents does not significantly affect the emergence of bias. This follows naturally from the design of our framework, in which each agent is initialized with a stance via a system prompt and generates responses based on the evolving conversation history. While increasing the number of agents lengthens this history, the bias phenomenon remains observable even in short interactions, as shown in Fig~\ref{fig:conversation_length}. This supports the generalizability of the findings from dyadic to multi-agent settings.}

\begin{table}[]
    \centering
    \caption{\added{Context window size for each tested model, expressed in number of tokens.}}
    
    \begin{tabular}{lc}
        \toprule
        Model ($L$) & Context Window Size ($W^L$) \\
        \midrule
        Gemini1.5 & 2M \\
  Claude3.5 & 200K \\
 Gpt4o & 128K \\
 Llama3.1 & 128K \\
 Qwen2.5 & 128K \\
 Zephyr & 33K \\
 Mixtral & 32K \\
 Gpt3.5 & 16K \\
 Gemma1.1 & 8K \\
 \bottomrule
    \end{tabular}
    \label{tab:context_window_size}
\end{table}
\textit{{\added{Memory loss}}.} {\added{Next, we investigate whether the observed bias can be due to memory loss. Specifically, we verify the saturation level of the context window of the given model $L$, which we denote as $W^L$. Such information is publicly available for the tested models, and is reported in Table~\ref{tab:context_window_size}. To assess if saturation occurred, we consider all the $50$ generated conversations for each combination of agent and topic. We further tokenize each conversation to assess the longest produced by a given model across all the topics. For each model $L$, we denote the number of tokens of such longest conversation as $T^L_{\textrm{max}}$. To ensure robustness, we perform the tokenization task by using two different tools: a BERT-based model tokenizer from HuggingFace ({\url{https://huggingface.co/docs/transformers/main_classes/tokenizer}}) and \texttt{tiktoken}, a crafted GPT-tokenizer released by OpenAI ({\url{https://pypi.org/project/tiktoken/}}). If the amount of estimated tokens differs, we consider the highest. To assess if a memory loss occurred, we therefore compute $T^L_{\textrm{max}}/W^L$, representing the portion of memory actually occupied by the longest conversation ever produced. The results are provided in Fig~\ref{fig:max_tokens_over_context_window_size}.
As shown, the highest percentage of memory ever occupied is about 64\%, when Zephyr is instantiated as agent, followed by Mixtral (30\%). We hypothesized this is due to the relatively small window context capacity of these models (see Table~\ref{tab:context_window_size}) combined with their average amount of tokens produced, displayed in Fig~\ref{fig:number_of_tokens}. 
Notably, this analysis shows that no agents display memory loss during the simulated conversations, thus excluding this factor as cause of the observed bias.
} 
}

\begin{figure}[t!]
    \centering
    \includegraphics[width=0.8\linewidth]{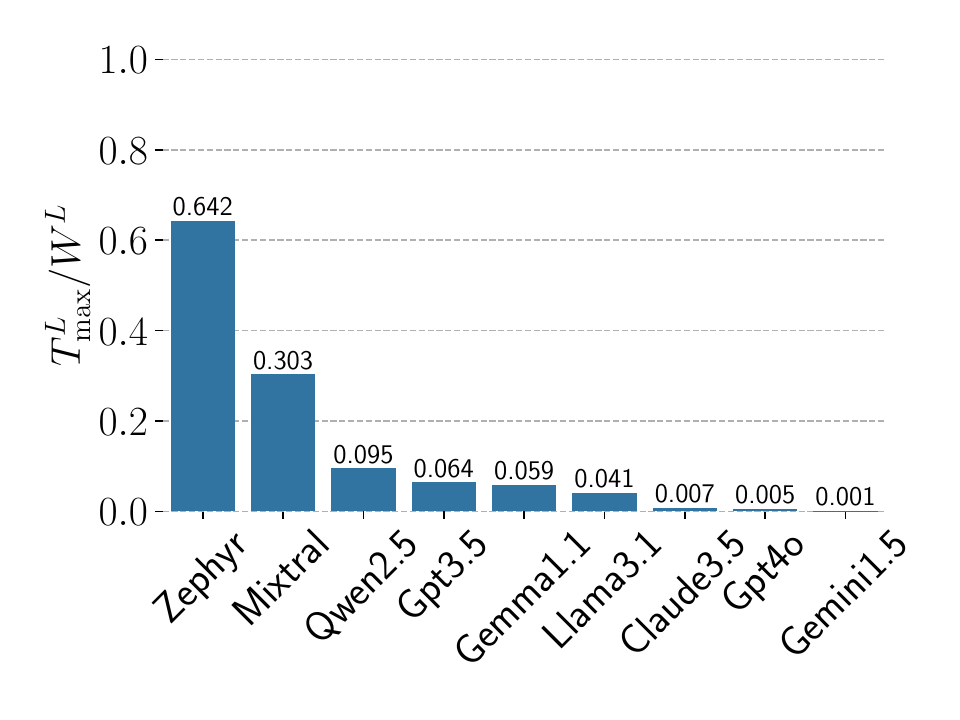}
    \caption{\added{Portion of context memory occupied by the longest conversation ever produced by a given model across all topics.}}
    \label{fig:max_tokens_over_context_window_size}
\end{figure}

\begin{figure}
    \centering
    \includegraphics[width=0.8\linewidth]{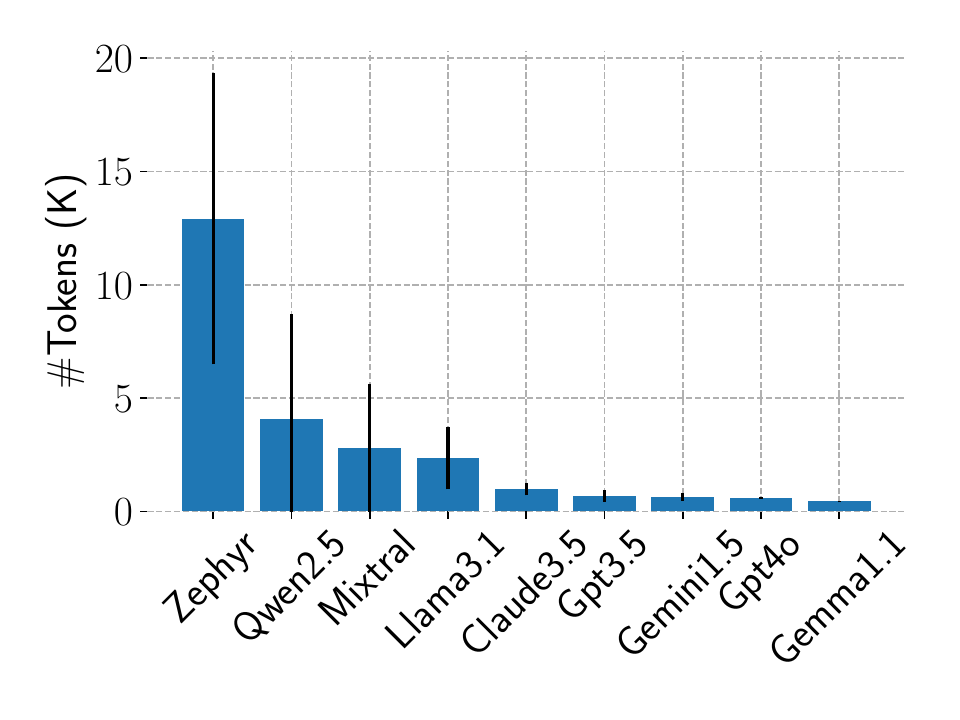}
    \caption{\added{Average number of tokens, expressed in thousands (K), generated by each model across all topics. Error bars represent standard deviations.}}
    \label{fig:number_of_tokens}
\end{figure}

\textit{{\added{Opinion signal agent.}}} \added{Finally, we assess the robustness of our results by varying the underlying opinion signal agent. Specifically, we adopted \texttt{Gemini-2.0-flash} ({\url{https://blog.google/technology/google-deepmind/google-gemini-ai-update-december-2024/}}), released by Google in 2024 with improvements in multimodal reasoning and long-context understanding. 
We use the conversation generated by Mixtral on ``Climate Change'', both prompted as Liberal and Conservative, as a representative sample. For a fair comparison, we prompted the opinion signal agent with the same protocol reported in the paragraph ``Opinion Agents''.}

\begin{table}[!ht]
    \centering
        \caption{\added{Percentage of conversations where an opinion drift was detected by any of the two opinion signal agents. The first (resp. second) row refers to conversations where conservative- (resp. liberal-) agents interacted. The last row depicts the aggregated results across all conversations. The column ``Agreement'' reports the percentage of conversations where both the opinion signal agents agree in terms of detected drift.}}
    \resizebox{\columnwidth}{!}{
    \begin{tabular}{cccc}
        \toprule
        Conversations w drift (\%) & LLaMa3-70B & Gemini-2.0-Flash & Agreement \\
        \midrule 
        Towards Liberal & 92 & 93 & 98 \\
        Towards Conservative & 0 & 30 & 70 \\
        \cmidrule(l){2-4}
        Total & 46 & 62 & 84 \\
        \bottomrule
    \end{tabular}
    }

    \label{tab:opinion_agents_comparison}
\end{table}
\added{The results are reported in Table~\ref{tab:opinion_agents_comparison}. The first (respectively, second) row shows the percentage of conversations with conservative- (respectively, liberal-) prompted agents in which a drift toward the Liberal (respectively, Conservative) pole was detected. The last row presents the aggregated results across all conversations. The column ``Agreement'' reports the percentage of conversations where the opinion signal agents agree in terms of drift (detected/non-detected).}

\added{Two key observations emerge. First, the two opinion-signal agents demonstrate an \textit{84\% agreement} across the 100 selected conversations (50 per political leaning). Second, Gemini-2.0-Flash is \textit{more likely} than LLaMa-3-70B to flag conversational drifts, labeling a larger portion of content as shifting toward the opposite pole.
These patterns imply that the results reported in the previous sections remain valid even under a majority-voting scheme, since the conversations where both agents agree coincide with those flagged by LLaMa-3-70B. In other words, LLaMa-3-70B effectively provides a lower-bound estimate of the proportion of simulations exhibiting an unwarranted opinion drift.}

\section*{Discussion}
\label{sec:discussion}

{The echo chamber setting in our framework provides a foundational benchmark for evaluating conversational bias. It exhibits both \textit{construct validity}, by offering a quantifiable outcome directly aligned with the theoretical definition of bias, and \textit{ecological validity}, by reflecting real-world contexts in which like-minded individuals frequently interact and reinforce shared beliefs. Additionally, it captures dynamics relevant to AI-coordinated (mis)information campaigns, where agents of the same stance collaborate~\cite{kanakaris2025networkinformedpromptengineeringorganized}.
Rather than encompassing all possible conversation types, our benchmark isolates a necessary condition for unbiased behavior. This is consistent with traditional bias assessments that rely on targeted questions, but our approach is more representative of naturalistic interaction patterns.}

{To enhance the robustness of our experimental evaluation, we tested several models from distinct LLM families, all of which have undergone fine-tuning with human feedback (RLHF). While examining pre-trained, non-fine-tuned models could offer insights into the origin of bias, we emphasize that (i) most state-of-the-art publicly available models are released only after RLHF; and (ii) prior research suggests that RLHF has limited influence on mitigating social biases~\cite{trainingllms}.}

Regarding the empirical evidence provided, a comparison of the left and right side of Fig~\ref{fig:heatmap-delta-bias} reveals that even LLMs exhibiting minimal or no bias under \added{direct probing} can display significant opinion shifts when engaged in interactive, multi-turn conversations. These findings underscore that conversational bias is not easily detectable and that simplistic testing approaches may fail to reveal deeper behavioral tendencies.

Notably, our results indicate that the manifestation of bias is neither uniform across topics nor consistent across models. A given agent may express little to no bias on one topic while exhibiting substantial opinion drift on another. Likewise, for the same topic, agents built on different LLMs may show varying degrees and directions of bias. This highlights the inherently complex and context-sensitive nature of bias in LLM-based systems, where both topic content and model architecture can strongly influence observed behavior.

Consequently, we contend that current methodologies for detecting bias in generated text are insufficient for auditing the nuanced behavior of LLMs in socially interactive settings. Even in structured echo chambers, where interactions are ostensibly homogeneous, we observe unexpected and unwarranted shifts in opinion, suggesting that more sophisticated evaluation frameworks are needed.

\section*{Conclusions and future work}
\label{sec:conclusions}

Large Language Models (LLMs) inherit social biases from their training data, which can persist or even worsen in their generated outputs. While existing bias detection methods, such as questionnaires and situational tests, identify biases in isolated settings, they fail to capture how LLMs behave in real-world multiagent interactions. 

The framework proposed in this paper is aimed at detecting conversational biases in multiagent LLM systems. By simulating chatroom debates with AI agents initially holding strong opinions, we observe unexpected opinion shifts, especially in conservative echo chambers, suggesting a latent liberal bias in many models. Notably, this bias is undetectable using conventional bias evaluation techniques. The findings highlight the need for more advanced, context-aware bias detection and mitigation strategies.

Given these considerations, it is also important to examine whether such unwarranted behavior extends beyond political opinion to other contexts. For instance, in discussions related to equity, law enforcement, or governance, it is crucial to determine whether LLM agents deviate from a neutral or balanced perspective without clear justification, potentially due to intrinsic biases.
Consider, for example, an LLM assistant used to analyze the resumes of job candidates and later engaged in a conversation about their strengths and weaknesses. If an inherent bias influences its responses, the resulting assessments could lead to unfair decision-making.

A consequential follow-up of this study is to investigate \emph{why} such specific biases emerge in LLMs, and to devise mitigation approaches. \added{A more in-depth analysis of conversational patterns could reveal how oscillations in the generated text of one agent may draw other agents towards strong attractor words that would then result into an opinion shift. Once an opinion shift occurs, other known behavioral biases of LLMs could contribute to a collective opinion shift. For example, LLMs are known to exhibit sycophancy~\cite{cheng2025elephant}, the tendency to agree with the text that is presented to them, even at the cost of correctness or consistency.} Notably, sycophancy could justify the unwarranted opinion change of the agent, \textit{consequently} to the opinion drift of their interlocutor, but does not explain why the agent did exhibit the drift itself. Indeed, as shown in the left side of Fig~\ref{fig:main-no_agents_changing_opinion}, in several configurations, only one agent exhibits conversational bias on average, thus leading us to discard sycophancy as the root cause of the observed opinion drifts. 
Nevertheless, even if sycophancy cannot explain the first drift during the conversation, it may be a phenomenon that contributes to the unwarranted opinion changes exhibited by the agents in subsequent steps of the conversation. {In this sense, we perform a complementary analysis where we compute the conditional probability of a second agent exhibiting a opinion change, following the first who already drifted (right side of Fig~\ref{fig:main-no_agents_changing_opinion}).}

As an additional starting point for future research, the specific generative frameworks underlying the models may not be robust to the exposure of the bias. 
This suggests that, besides solely evaluating the biases through the generated text, future work can look \emph{inside} the models and explore how the probabilities over the output tokens evolve throughout the discussion concerning the emerging bias. For instance, we can conduct an in-processing mitigation strategy by refining the weights of the model, thus making its responses insensitive to the bias.



\section*{Limitations}
{This work presents several limitations that open avenues for future research and refinement. 
\begin{enumerate}
    \item The study centers primarily on polarizing U.S. political topics, such as climate change and abortion~\cite{gallup2023partisan, pew2023inflation}. While these are salient and well-documented issues, this focus may limit the generalizability of findings to less contentious or non-U.S. contexts, where the structure of opinion formation and political engagement may differ significantly.
\item The reliance on an automated stance detection system introduces an additional layer of abstraction and complexity. Although the used model proves to accurately perform the task, it may introduce potential inaccuracies and biases inherent to the stance classifier itself. An alternative approach could involve endowing conversational agents with the ability to explicitly signal their positions, increasing interpretability and reducing dependence on post-hoc classification tools.
\item The scope of the work is currently limited to simulations of opinion dynamics within multiagent interactions. While valuable, this narrow task focus may not capture the full spectrum of challenges in human-AI communication. Future work should explore how similar dynamics unfold in other collaborative or adversarial multiagent scenarios, such as joint decision-making, negotiation, or misinformation detection.
\item The range of agent roles modeled in the simulations is relatively constrained. Agents are treated largely as homogeneous participants, lacking more nuanced behavioral archetypes (e.g., skeptics, persuaders, or contrarians). Incorporating diverse agent personas could lead to richer dynamics and a better approximation of real-world conversational ecosystems.
\item The framework has been tested on a comprehensive but limited set of underlying language models. Expanding the evaluation to include a broader range of models, including those with varying capacities, training data, and architectural inductive biases, could provide deeper insights into the generalizability and robustness of the observed phenomena.
\end{enumerate}
}

\section*{Acknowledgments}

We thank Simon Martin Breum, Daniel Vædele Egdal, Victor Gram Mortensen, and Anders Giovanni Møller for
providing the computational framework to simulate agent conversations, and for the early research insights of their
work that motivated our contribution. 


%
%

\appendix

\section*{Appendix}
\begin{figure}[!ht]
    \centering
    \includegraphics[width=0.8\linewidth]{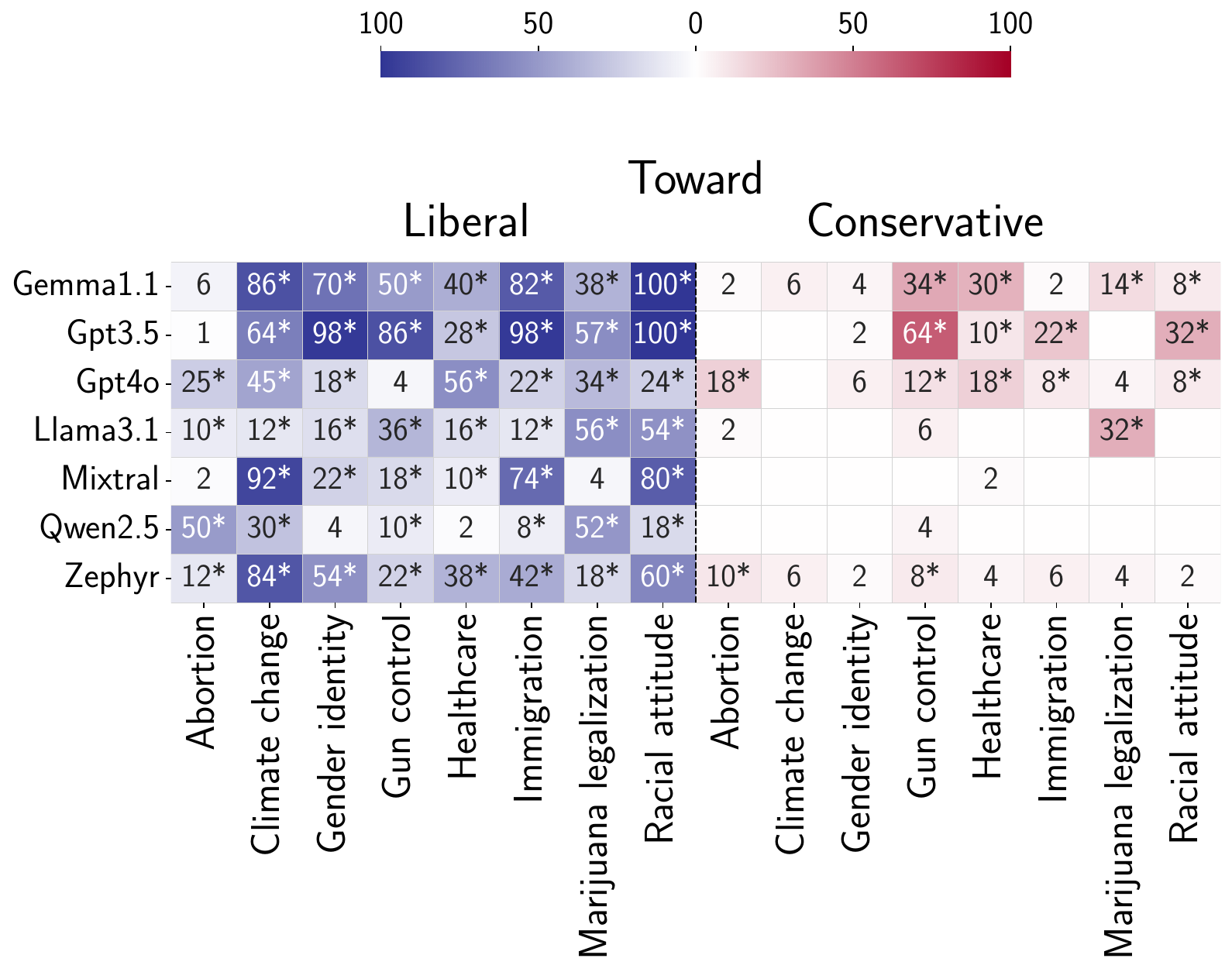}
    \caption{Percentage of chatroom simulations where at least one unwarranted opinion change occurs. The ``*" symbol indicates statistical significance with $p < 0.05$ under the Z-test~\cite{casella2024statistical}. Empty cells indicate 0-values, while cells with the "-" symbol indicate unavailable results. The darker the color of the cell, the more prominent the detected shift.}
    \label{fig:heatmap-bias}
\end{figure}
We provide here additional results, including: the percentage of conversations in which at least one unwarranted opinion change occurs (Fig~\ref{fig:heatmap-bias}); drift toward neutral positions (Fig~\ref{fig:appendix-bias}); detailed information on how many agents altered their opinions during the chat (Fig~\ref{fig:heatmap-no-agents}); and finally, on when the unwarranted opinion drifts happen, in terms of conversation length $M$ (Fig~\ref{fig:single_conversation_length}).

Starting with the former, we show that most of the observed drifts are statistically significant (marked with a ``*''). 
In particular, for each estimated proportion, we compute a 95\% confidence interval using a normal approximation to the binomial distribution (with $n = 50$), and we flag it as statistically significant whenever the lower bound of the interval is strictly greater than zero. Going further in details, Fig~\ref{fig:appendix-bias} reveals that, in most cases, we capture not only shifts toward a neutral stance, but a substantial number of conversations exhibit a pronounced drift, particularly toward left-leaning positions. This suggests that the observed changes are not merely neutralizing tendencies but instead directional shifts that may indicate systematic influence.
Similarly, Fig~\ref{fig:heatmap-no-agents} demonstrates that, across the majority of configurations:
(i) opinion changes predominantly trend toward a Liberal orientation, and
(ii) in some instances, both agents exhibit a shift. Notable examples include Gemma 1.1 and GPT-3.5 on the topic of Racial Attitude, as well as GPT-4o and Zephyr in discussions about Climate Change. These patterns further confirm the findings previously discussed, reinforcing the evidence of political bias in agent interactions.
In both Figs, we remark that the results with Claude3.5 towards the Liberal pole are not reported since the agent refused to impersonate the Conservative persona, as discussed in Section ``Experimental setup".

\begin{figure*}[!ht]
    \centering
    \includegraphics[width=\linewidth]{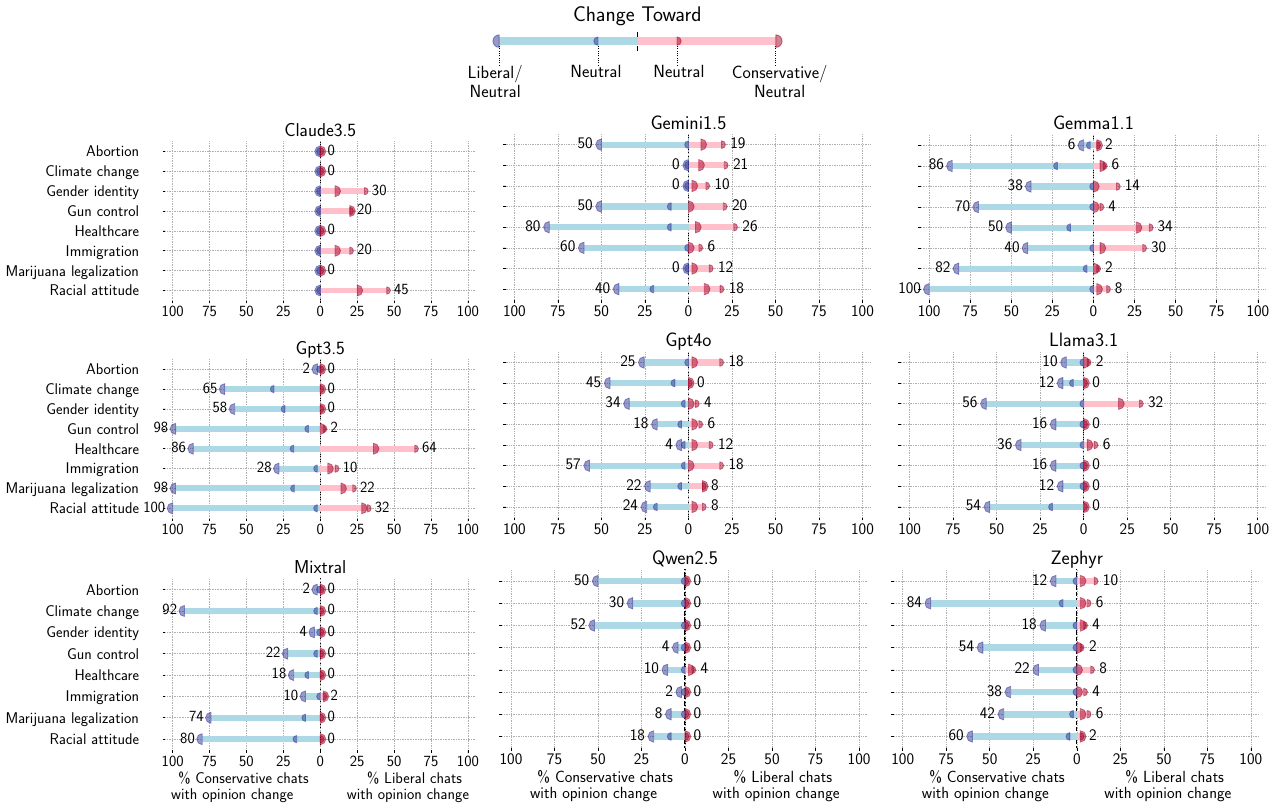}
    \caption{Percentage of chatroom simulations where at least one unwarranted opinion change occurs, indicating drifts also towards neutral positions.}
    \label{fig:appendix-bias}
\end{figure*}

\begin{figure*}[!ht]
    \centering
    \includegraphics[width=\linewidth]{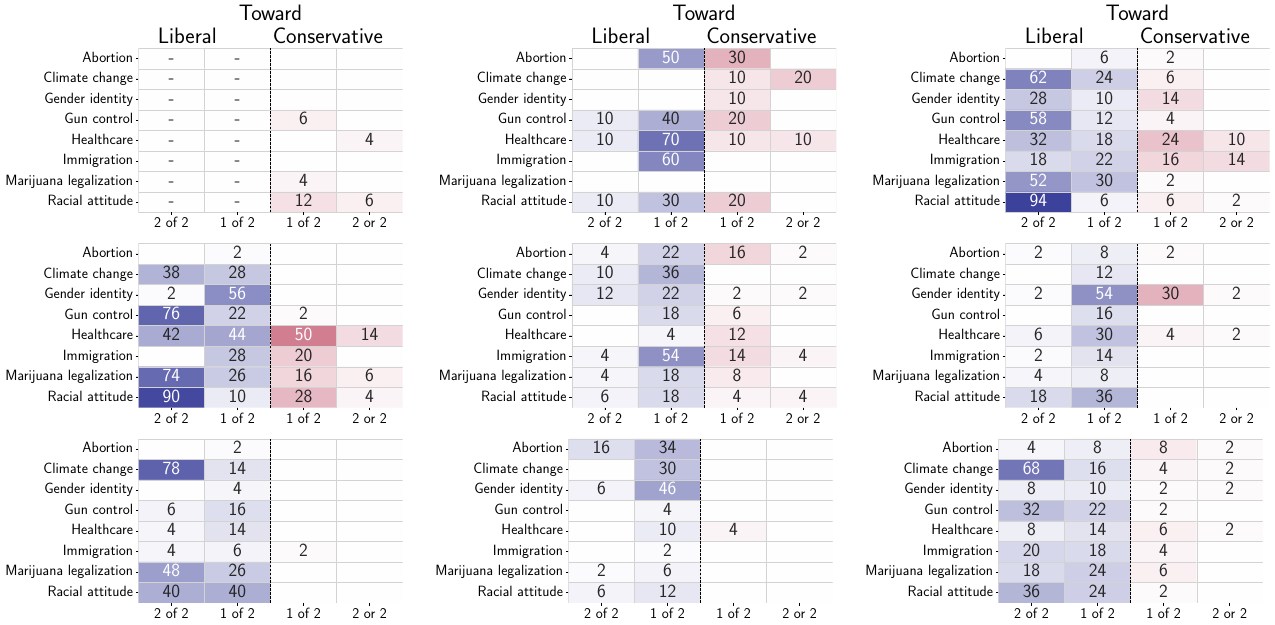}
    \caption{Percentage of chatroom indicating whether one or both agents changed opinion during the conversation. The darker the color, the higher the value. Empty cells represent 0-scores, while cells with the "-" symbol indicate unavailable results.  In order, from top-left to bottom-right: Claude3.5, Gemini1.5, Gemma1.1, GPT3.5, GPT4o, LLaMa3.1, Mixtral, Qwen2.5, and Zephyr.}
    \label{fig:heatmap-no-agents}
\end{figure*}

\begin{figure}[!ht]
    \centering
    \includegraphics[width=\columnwidth]{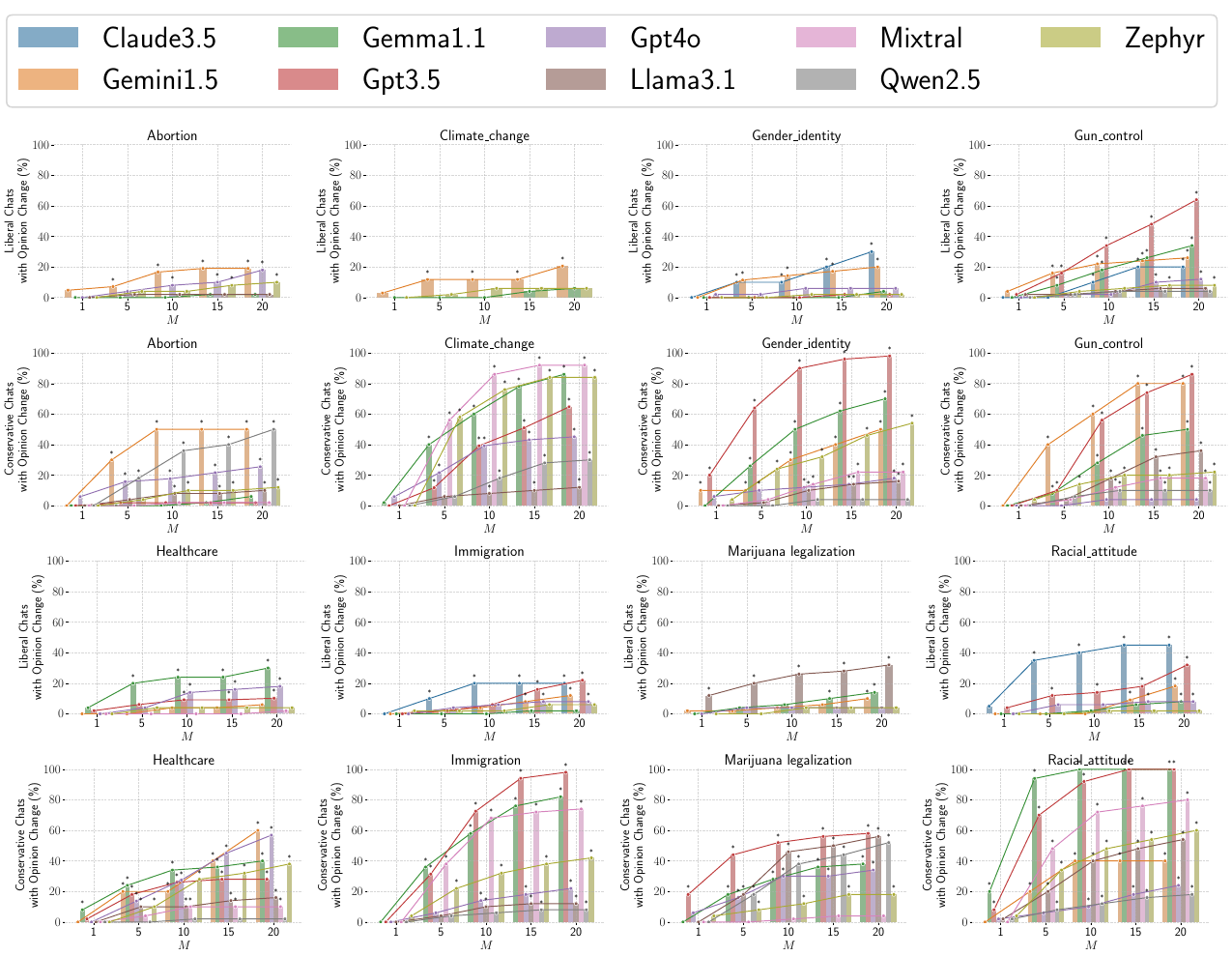}
    \caption{{For each selected topic, percentage of Liberal (upper row) and Conservative (lower row) conversations exhibiting an unwarranted opinion change (Y-axis) by varying the conversation length $M$ (X-axis). The symbol ``*'' above bars denotes statistical significance.}}
    \label{fig:single_conversation_length}
\end{figure}
\begin{figure}[!ht]
    \centering
     \includegraphics[width=\linewidth]{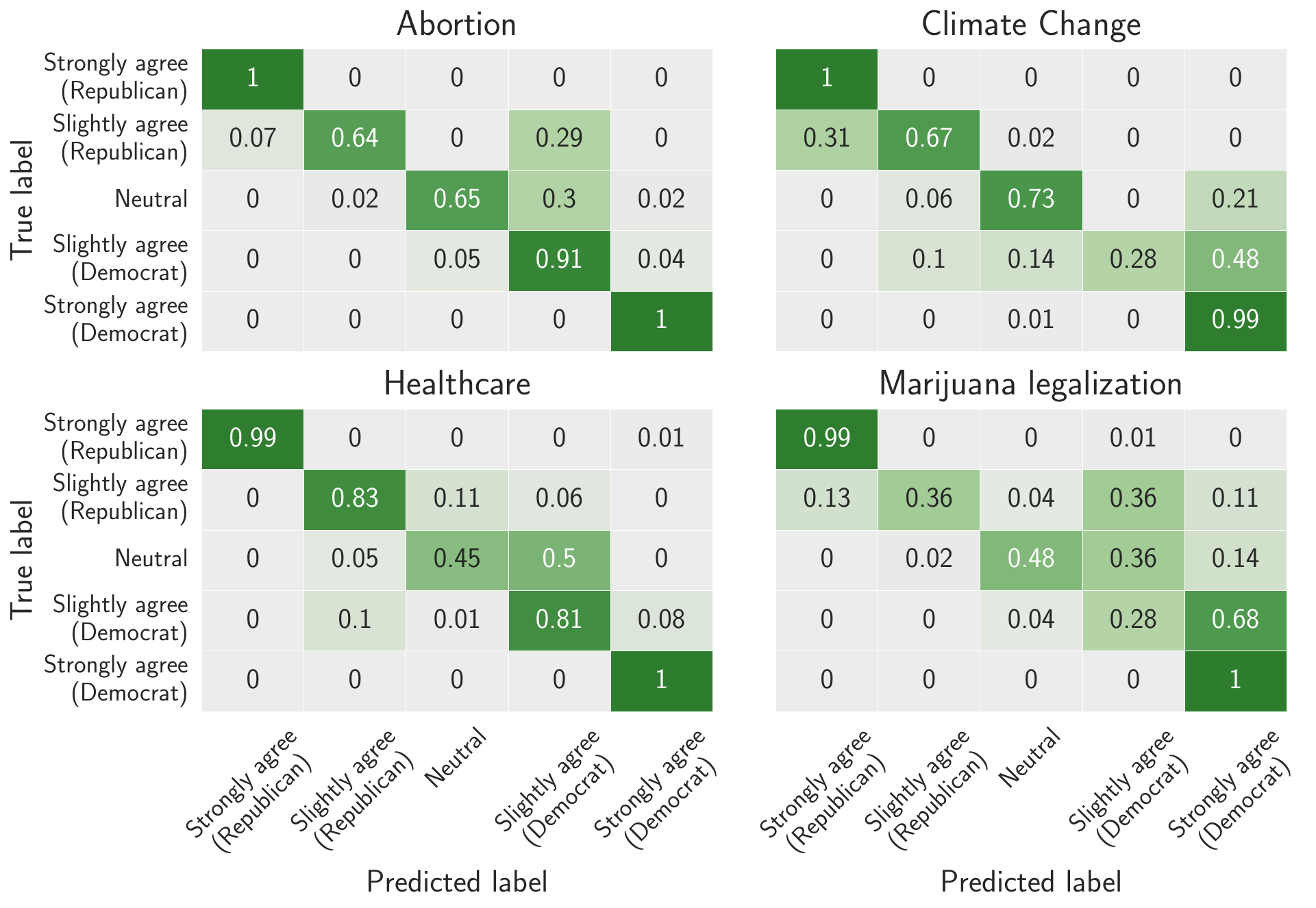}
    \caption{Confusion Matrix of LLaMa-3-70B on the topics of Abortion, Climate Change, Healthcare, and Marijuana Legalization. The results show that the opinion signal agent is accurate over both the Liberal and the Conservative pole, especially on strong agreement.}
    \label{fig:confusion-matrix}
\end{figure}

{Further, Fig~\ref{fig:single_conversation_length} shows the percentage of Liberal/Conservative conversations with an unwarranted opinion drift at a topic-level, estimated as $M$ varies in the range [1, 5, 10, 15, 20]. The asterisk (``*'') denotes statistical significance. Notably, we show that most models exhibit drift in the very first steps of the conversation, especially when they are instantiated as conservative agents (lower row).}

{Finally, we report the confusion matrix of LLaMa-3-70B as opinion signal agent, showing that the model displays high accuracy over both the Liberal and the Conservative pole (especially on strong agreement).}

\bibliography{ref}

\end{document}

%% file: macros.tex
\newcommand{\noagents}{\ensuremath{N}\xspace}
\newcommand{\nomessages}{\ensuremath{M}\xspace}
\newcommand{\timestep}{\ensuremath{t}\xspace}

\newcommand{\added}[1]{\textcolor{black}{#1}}

\newcommandx{\unsure}[2][1=]{\todo[linecolor=red,backgroundcolor=red!25,bordercolor=red,caption={},#1]{#2}}
\newcommandx{\change}[2][1=]{\todo[linecolor=blue,backgroundcolor=blue!25,bordercolor=blue,caption={},#1]{#2}}
\newcommandx{\info}[2][1=]{\todo[linecolor=green,backgroundcolor=green!25,bordercolor=green,caption={},#1]{#2}}
\newcommandx{\improvement}[2][1=]{\todo[linecolor=purple,backgroundcolor=purple!25,bordercolor=purple,caption={},#1]{#2}}
\newcommandx{\discussion}[2][1=]{\todo[linecolor=blue,backgroundcolor=yellow!25,bordercolor=yellow,caption={},#1]{#2}}

\newcommandx{\thiswillnotshow}[2][1=]{\todo[disable,#1]{#2}}